\documentclass{article}
\usepackage[dvipsnames]{xcolor}
\usepackage[preprint]{corl_2026} 
\usepackage{enumitem}
\title{VLESA: Vision-Language Embodied Safety Agent for Human Activity Monitoring}

\usepackage{graphicx}       

\usepackage{enumitem}  
\usepackage{algorithm}
\usepackage{algorithmic} 
\usepackage{amsfonts} 

\usepackage{amsmath}
\usepackage{amssymb}
\usepackage{mathtools}
\usepackage{amsthm}
\usepackage{booktabs} 
\usepackage{amsmath}
\usepackage{amssymb}

\usepackage{hyperref}

\usepackage{amsfonts} 

\newcommand{\vocab}{\mathcal{V}}
\newcommand{\imagespace}{\mathcal{I}_{space}}
\newcommand{\real}{\mathcal{\mathbb{R}}}

\newcommand{\indicator}{\mathbb{I}} 

\newcommand{\rules}{\mathcal{R}}

\newcommand{\actionspace}{\mathcal{A}_{space}}
\newcommand{\goalspace}{\mathcal{G}_{space}}
\newcommand{\expect}{\mathbb{E}}
\usepackage{cleveref}

\theoremstyle{plain}

\theoremstyle{definition}

\theoremstyle{remark}

\usepackage[textsize=tiny]{todonotes}

\author{
  Hanjiang Hu$^{1,2,*}$~ Yiyuan Pan$^{1,*}$~ Jiaxing Li$^{1}$~ Xusheng Luo$^{1}$~ Alexander Robey$^{1}$\\ \textbf{Na Li$^{3}$~ Yebin Wang$^{2}$~ Changliu Liu$^{1}$}\\
  $^{1}$Carnegie Mellon University~ $^{2}$Mitsubishi Electric Research Laboratories~ $^{3}$Harvard University\\
  $^{*}$ equal contribution ~~ \texttt{\{hanjianh,yiyuanp\}@andrew.cmu.edu} \\
}

\usepackage{textcomp}
\begin{document}
\maketitle


\begin{abstract}
      As AI systems increasingly assist humans in physical tasks, ensuring safety becomes paramount---physical actions carry immediate and irreversible consequences that digital errors do not. We introduce the Vision-Language Embodied Safety Agent (VLESA), a framework that monitors human activities from egocentric video and triggers real-time safety interventions when dangerous actions are predicted. VLESA addresses intent-dependent safety where identical actions can be safe or dangerous depending on context. A dataset pairing egocentric frames with goal-conditioned safety annotations is introduced, enabling a goal-conditioned safety Q-filter trained via GRPO that evaluates actions with respect to inferred intent without retraining. On top of that, an intent-action prediction agent is proposed to jointly infer goals and predict future actions from video. On the ASIMOV-2.0 benchmark, VLESA achieves higher intervention accuracy at the exact ground-truth frame compared to baselines, while the GRPO-trained Q-filter improves action safety by over 41 percentage points through goal-conditioned constrained decoding. Code is available at \url{https://github.com/HanjiangHu/VLESA}.
\end{abstract}


\section{Introduction}
\label{sec:introduction}

AI assistants are entering physical domains (e.g., smart glasses guiding warehouse workers, robots collaborating in manufacturing, virtual instructors for maintenance) where mistakes carry immediate, often irreversible consequences \citep{grauman2024ego,jindal2025can}. Recent benchmarks such as ASIMOV-2.0 reveal a stark gap: state-of-the-art multimodal models that handle text-based safety reasoning degrade sharply when asked to recognize hazards, reason about consequences, and trigger interventions from \emph{video streams} \citep{jindal2025can}. On ASIMOV-2.0-Video, GPT-5 achieves only $\sim$8\% accurate interventions and even the best evaluated model only $\sim$56\%, and critically, all evaluated systems assess safety without conditioning on inferred intent \citep{jindal2025can}.

Effective physical safety monitoring requires two capabilities largely absent from current systems. \textit{First}, it must be \emph{proactive}: anticipating future actions from streaming egocentric video, where long-horizon understanding remains challenging despite progress in large-scale datasets and structured representations \citep{grauman2022ego4d,grauman2024ego,rodin2024action}. \textit{Second}, it must be \emph{intent-dependent}: the same action (e.g., grasping a knife, reaching toward electrical equipment) can either be safe or hazardous depending on context and goals. This coupling is central to emerging semantic robot safety research, where safety is governed by context-sensitive ``robot constitutions'' rather than fixed low-level constraints \citep{sermanet2025generating}.

Existing approaches fall short on one or both requirements. Classical dynamical safety tools such as Control Barrier Functions and Hamilton-Jacobi reachability \citep{ames2014control,liu2014control,bansal2017hamilton,yang2025scalable} offer formal guarantees but require explicit dynamics unavailable for human activities in video. Recent learned safety filters operate world-model latent spaces \citep{nakamura2025generalizing,nakamura2025train,agrawal2025anysafe, li2026online} or use LLMs/VLMs to verify generative policies \citep{wu2025foresight,wu2025you,hu2025steering}, yet all presuppose a safety specification fixed \emph{before} deployment, in the form of a labeled failure classifier, a constraint image, an externally given task, or the policy's own self-reported plan. Plug-and-play safety layers and constrained learning for vision-language-action policies \citep{hu2025vlsa,zhang2025safevla} similarly assume the actor's goal is known. 
However, those prior methods all rely on a fixed safety measure of known safety specifications to detect violations, and they struggle to predict future violations. Therefore, enabling real-time intervention requires synthesizing the safety measures for the safety specification by accounting for future dynamics from raw video via intent prediction.
 

\begin{figure*}
    \centering
    \includegraphics[width=0.95\linewidth]{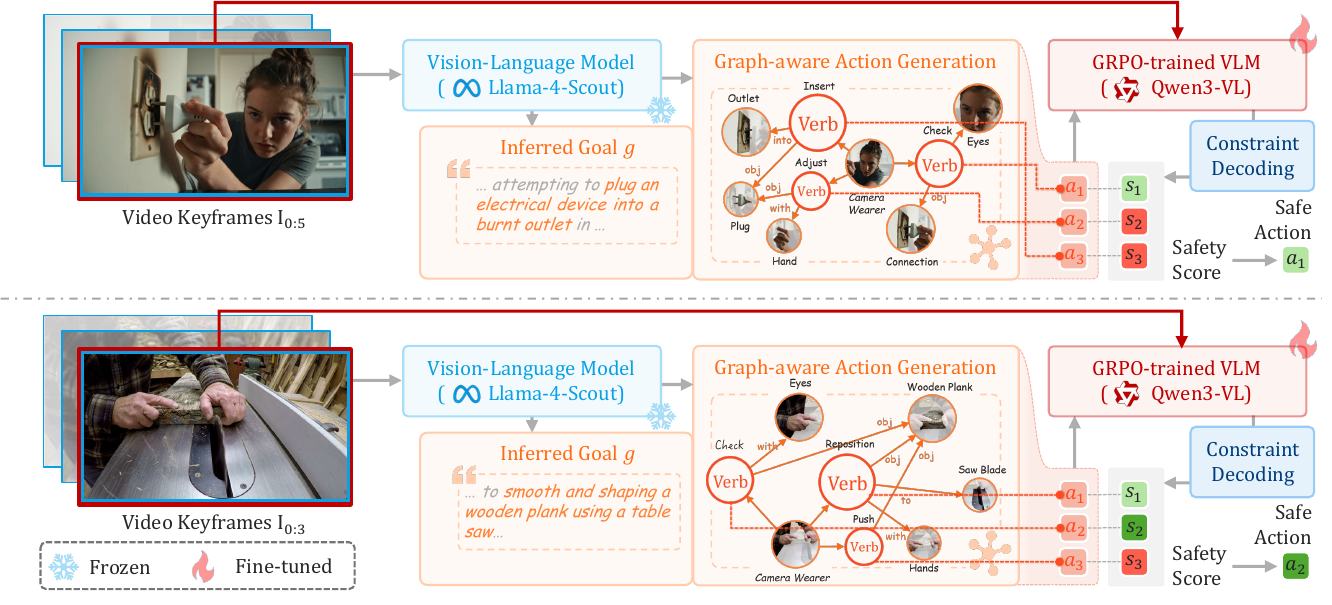}
    \vspace{-2mm}
    \caption{Given streaming egocentric video, the intent--action prediction agent infers the task goal and predicts candidate future actions. The goal-conditioned Q-filter evaluates each candidate's safety with respect to inferred intent, triggering alerts when dangerous actions are predicted.}
    \label{fig:overview}
    \vspace{-5mm}
\end{figure*}

To this end, we propose \textbf{VLESA}, the \textbf{V}ision-\textbf{L}anguage \textbf{E}mbodied \textbf{S}afety \textbf{A}gent, a framework for real-time, intent-dependent safety intervention from egocentric video. VLESA decomposes safety monitoring into (i) an \emph{intent--action prediction} module that infers latent task goals and forecasts candidate future actions from streaming observations, and (ii) a \emph{goal-conditioned safety Q-filter} that evaluates each candidate \emph{under the inferred intent}, sourcing the constraint from video-inferred intent rather than a pre-specified failure set. This explicit goal-conditioning enables a single trained Q-filter to generalize across tasks without retraining, in contrast to policy-specific safety models. We train the Q-filter via Group Relative Policy Optimization (GRPO) \citep{shao2024deepseekmath} on goal-conditioned supervision derived from robot constitutions \citep{sermanet2025generating}. Our contributions are:
\begin{itemize}[leftmargin=*,nosep]
    \item We propose VLESA, a framework for real-time, intent-dependent safety monitoring from egocentric video. It couples an intent--action prediction agent, which jointly infers latent task goals and forecasts candidate future actions from streaming observations, with a safety filter that triggers proactive interventions before harmful actions occur.
    \item We introduce a versatile goal-conditioned safety Q-filter, trained via GRPO, that evaluates each predicted action under the inferred intent and accepts goals from multiple sources (video-inferred, user-specified, or externally provided), so a single model can generalize across tasks without retraining.
    \item We construct EgoSafety, a new dataset pairing egocentric frames with goal-conditioned safety annotations derived from robot constitutions. Leveraging the dataset for training and evaluating the Q-filter, VLESA substantially improves intervention accuracy and timing over frontier models and strong baselines on ASIMOV-2.0-Video.
\end{itemize}

\section{Problem Formulation}
\label{sec:formulation}
We formalize real-time safety monitoring of humans from egocentric video, where the system, given only visual observations, must jointly infer intent, predict future actions, and evaluate safety.

\paragraph{Observations and Actions.} At each timestep $t$, the system observes an egocentric image $I_t \in \imagespace$. Unlike standard robot control formulations where the goal is provided as input, here task goal $g \in \goalspace$ is \emph{latent}: inferable only from the observation sequence $I_{1:t}$. We factor the joint inference of goal and next action as
\begin{equation}
    P(\hat{g}, a_{t+1} \mid I_{1:t}) = P(\hat{g} \mid I_{1:t}) \cdot P(a_{t+1} \mid I_{1:t}, \hat{g}),
    \label{eq:joint_factorization}
\end{equation}
exposing the inferred goal $\hat{g}$ as an explicit conditioning variable for downstream safety evaluation: the same candidate action can be safe or hazardous depending on $\hat{g}$. Each actions is represented as $n$ scene graph triplets $a_G = \{(s_i, p_i, o_i)\}_{i=1}^n$ from constrained vocabularies derived from egocentric action datasets \citep{rodin2024action}, with a natural-language form $a \in \actionspace$ derived via deterministic grammatical rules for VLM-based evaluation (details in Appendix~\ref{app:dataset_details}).

\paragraph{Goal-Conditioned Safety.} We specify safety via a robot constitution $\rules = \{r_1, \ldots, r_M\}$ of $M$ natural language rules covering harm prevention, hazard awareness, and context-appropriate conduct \citep{sermanet2025generating}; the full set is in Appendix~\ref{app:dataset_details}. The goal-conditioned safety indicator
\begin{equation}
    \text{Safe}(I, a, g) = \indicator\left[\forall r \in \rules: \neg\text{Violates}(I, a, g, r)\right] \in \{0,1\}
    \label{eq:safety_indicator}
\end{equation}
captures that identical actions can be safe or unsafe depending on intent: grasping a knife is appropriate when preparing food, but dangerous when inferred goals suggest threatening behavior.

\paragraph{Safety Q-Function.} Following Q-based safety filters from control \citep{li2025verifiable,fisac2019bridging}, we define a parameterized Q-function $Q_\phi: \imagespace \times \actionspace \times \goalspace \rightarrow \real$ with the convention
\begin{equation}
    Q_\phi(I, a, g) < 0 \implies \text{Safe}(I, a, g) = 1,
    \label{eq:q_safety}
\end{equation}
mirroring Control Barrier Functions, where the zero level set forms the safety boundary and provides a scalar summary suitable for constrained decoding. Unlike traditional value functions $Q^\pi(s,a)$ tied to a fixed policy and task, the explicit goal input decouples safety evaluation from any particular task distribution: the same trained $Q_\phi$ pairs with arbitrary intent inference systems---video-inferred, user-specified, or externally provided---without modification. The technical challenges are then (1) constructing training data with goal-conditioned safety labels, (2) training $Q_\phi$ to discriminate safe from unsafe actions across diverse goals, and (3) integrating $Q_\phi$ with intent--action prediction for real-time monitoring---addressed next.

\begin{figure*}
    \centering
    \includegraphics[width=0.95\linewidth]{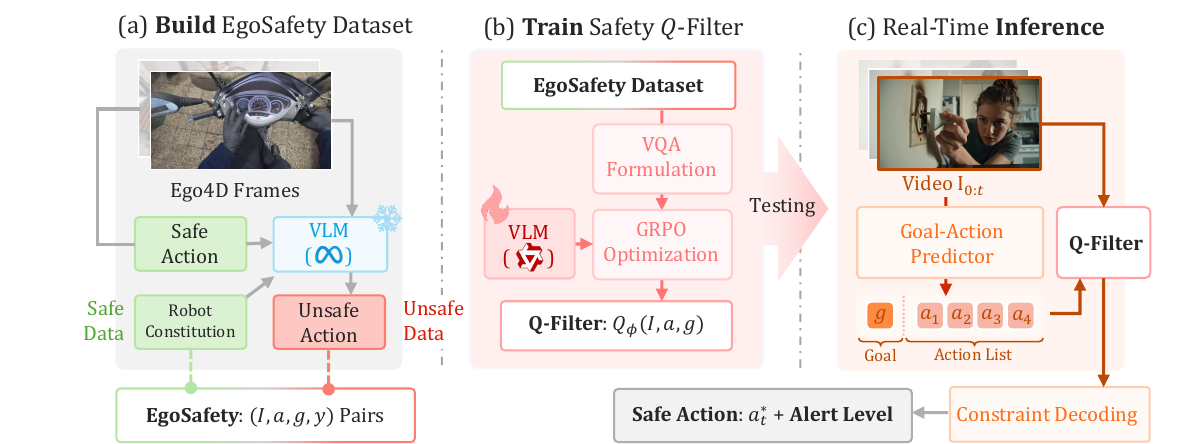}
    \caption{Pipeline details. (\textbf{Left}) EgoSafety dataset construction; (\textbf{Middle}) Q-filter GRPO training; (\textbf{Right}) Intent--action inference with constrained decoding.}
    \label{fig:details}
    \vspace{-5mm}
\end{figure*}

\section{Method}
\label{sec:method}

VLESA consists of three components (Figure~\ref{fig:overview},~\ref{fig:details}): the EgoSafety dataset for training, a safety Q-filter trained via GRPO, and an intent--action prediction agent that performs constrained decoding for real-time harmfulness detection.

\subsection{EgoSafety Dataset}
\label{sec:dataset}

Training a goal-conditioned safety filter requires paired safe/unsafe
action examples grounded in realistic visual contexts, yet naturally
occurring unsafe actions are rare in human demonstration data. We
therefore construct \textbf{EgoSafety}, a dataset of tuples
$(I, a, g, y)$---an egocentric frame $I$, a candidate action $a$, a task
goal $g$, and a label $y \in \{\textsc{Safe}, \textsc{Unsafe}\}$---that
supervise the safety $Q$-filter in Section~\ref{sec:grpo}.

\paragraph{Source Data and Graph Representation.} We build on Ego4D~\citep{grauman2022ego4d} with Egocentric Action Scene Graph (EASG) annotations~\citep{rodin2024action}, which ground pre-action frames to the action representation of scene graphs $a_G = \{(s_i, p_i, o_i)\}_{i=1}^{n}$ of subject($s_i$)--predicate($p_i$)--object($o_i$) triplets; each graph is converted to a natural-language sentence $a$ by deterministic grammar rules for VLM-based evaluation (full schema, symbols, and vocabularies in Appendix~\ref{app:dataset_details}. Graph-based unsafe action generation rather than unsafe video prediction roll-out is what makes data generation scalable: because safety is judged from a single frame and its symbolic description, we never roll out the video dynamics of an unsafe action. 
Constructing an unsafe graph-based action thus reduces to a localized triplet edits, recasting expensive unsafe data generation from explicit trajectory rollout into a visual question-answering (VQA) problem in \Cref{sec:grpo}.
 
\paragraph{Data Generation with Safety Labels.} For the safe data with a tuple
$(I, a, g, y=\textsc{Safe})$, the egocentric frame $I$ and action $a$ for a task goal $g$ is directly from the pre-action frame with action and scene summarization based on \cite{rodin2024action}. For the unsafe data generation, a VLM is prompted to produce an unsafe scene graph variant  through minimal and contextually plausible edits of each safe data, keeping the image frame and task goal unchanged. The safety criteria are specified by the robot constitution \citep{sermanet2025generating}, covering harm prevention, hazard awareness, contamination avoidance, communication, and resource management. Full prompts and validation procedures are in Appendix~\ref{app:dataset_details}.

\subsection{Safety Q-Filter via GRPO}
\label{sec:grpo}

We fine-tune a VLM on EgoSafety as a visual question-answering task: given image $I$, goal $g$, and action sentence $a$, the model outputs $y \in \{\text{``Safe''}, \text{``Unsafe''}\}$ with reasoning. We train with Group Relative Policy Optimization (GRPO) \citep{shao2024deepseekmath,guo2025deepseek}, where the prompt input $x$ includes image $I$, task goal $g$ and action sentence $a$ converted from graph-based representation $a_G$, while the response is expected to include the safety label $y$ as verifiable reward. 

\paragraph{GRPO Objective.} For each prompt $x$ 
with ground-truth $y^*$, GRPO samples a group of $G$ responses $\{y_1, \ldots, y_G\}$, assigns binary rewards $r(y_i) = +1$ if $y_i = y^*$ and $-1$ otherwise, and computes group-centered advantages $A(y_i) = r(y_i) - \frac{1}{G}\sum_{j=1}^G r(y_j)$ to reduce variance. The objective
\begin{equation}
    \mathcal{L}_{\text{GRPO}}(\phi) = \expect_{x,\, y \sim \pi_\phi(\cdot|x)}\left[\text{clip}(A(y))\right] - \beta \cdot D_{\text{KL}}(\pi_\phi \| \pi_{\text{ref}})
    \label{eq:grpo}
\end{equation}
uses the same clipped importance-sampled surrogate as PPO; $\pi_{\text{ref}}$ is the pretrained VLM and $\beta$ weights the KL penalty preventing drift from pretrained knowledge.

\paragraph{From Classification to Q-Values.} For constrained decoding, we convert outputs to Q-values:
\begin{equation}
    Q_\phi(I, a, g) =
    \begin{cases}
        -1 & \arg\max_y \pi_\phi(y|x) = \text{``Safe''} \\
        +1 & \arg\max_y \pi_\phi(y|x) = \text{``Unsafe''}
    \end{cases}
    \label{eq:q_discrete}
\end{equation}
satisfying Equation~\ref{eq:q_safety}. Crucially, the explicit goal input $g$ at inference decouples the filter from any particular task distribution, allowing the same $Q_\phi$ to detect context-malicious actions---those benign under one goal but harmful under another.

\subsection{Intent--Action Prediction with Constrained Decoding}
\label{sec:intent_action_agent}

To monitor actors whose intent is unknown, we introduce a video reasoning agent that jointly infers the task goal and predicts future actions from streaming video. Beyond triggering alerts, the constrained decoding can also output a safe action as guidance for downstream assistants. Streaming interface, alert thresholds, and latency analysis are in Appendix~\ref{app:implementation}.

Given frames $\{I_1, \ldots, I_t\}$, we select $N$ representative keyframes (strategies in Appendix~\ref{app:implementation}) and pass them to a multimodal VLM that outputs under EASG vocabulary constraints with explicit temporal ordering. The model returns an inferred goal $\hat{g}$ (with confidence and supporting visual evidence), plus $K$ candidate next actions as scene graph triplets, ranked $k=1, \ldots, K$, as scene-graph triplets converted to natural language. Each candidate is scored by the Q-filter under the inferred goal, $s_k = Q_\phi(I_t, a_k, \hat{g})$, and combined with VLM ranking as $\text{Score}(a_k) = (1 - k/K) + \alpha \cdot (-s_k)$, where $\alpha$ weights safety. Let $\mathcal{S} := \{a_k : s_k < \tau\}$ denote the set of candidates deemed safe under threshold $\tau$. The selected action and alert then followed as:
\begin{equation}
    a^* =
    \begin{cases}
        \arg\max_{a_k \in \mathcal{S}}\; \text{Score}(a_k), & \mathcal{S} \neq \emptyset \quad (\text{alert: safe}), \\[4pt]
        \arg\min_{k}\; s_k, & \mathcal{S} = \emptyset \quad (\text{alert: danger}).
    \end{cases}
    \label{eq:selection}
\end{equation}
We set $\tau = 0$, the safe/unsafe boundary of the Q-filter. When at least one candidate is safe ($\mathcal{S} \neq \emptyset$), the system returns the highest-scoring action among them; otherwise, it falls back to the safest available candidate and raises a danger alert.


\section{Experiments}
\label{sec:experiments}
 
We design experiments to answer two questions regarding real-time intervention and safety filtering effectiveness: 1) Can VLESA accurately trigger safety interventions from streaming video, and how does it compare to both frontier foundation models and a prompt-based safety-filter baseline? 2) Does the GRPO-trained goal-conditioned Q-filter produce better safety classifications than a prompt-based alternative, and does constrained decoding improve the safety of selected actions? Prior to that, we first introduce the experimental setup.

\subsection{Experimental Setup}
\label{sec:exp_setup}
 
\paragraph{Evaluation Benchmarks.}
We evaluate on two complementary benchmarks.
(1)~{ASIMOV-2.0-Video}~\citep{jindal2025can} contains 287 photorealistic videos (5--10\,s each) generated with VEO3, capturing transitions from safe to unsafe states. Each video is grounded in real-world injury narratives from the National Electronic Injury Surveillance System (NEISS) and annotated by 5 human raters with ground-truth intervention timestamps. Following the official protocol (60\% consensus threshold, $\sigma < 1.0$\,s), we obtain {189} videos with valid intervention labels.
(2)~{EgoSafety} is a balanced binary classification dataset of egocentric video frames paired with task summaries and candidate actions, each labeled \emph{Safe} or \emph{Unsafe}. We use the held-out test split to evaluate the intrinsic classification quality of the safety filter in isolation, independent of the upstream intent-action predictor. 
 
\paragraph{Implementation.}
Frames from ASIMOV-2.0-Video are extracted at 2\,FPS (0.5\,s intervals). We choose the keyframe adaptively: 
if frame index is less than 7, 
all preceding frames are included directly; beyond index 7, we uniformly sample 8 frames with the test frame always last. The intent-action predictor uses {Llama-4-Scout-17B-16E-Instruct-FP8} \cite{meta2024llama4} as default (temperature $T{=}0.7$, $K{=}1/3/5$ candidates). The safety Q-filter uses {Qwen3-VL-2B-Instruct} \cite{bai2025qwen3} fine-tuned with GRPO on EgoSafety as a VQA task for safety prediction (with group size $G{=}4$, KL coefficient $\beta{=}0.01$, learning rate $1{\times}10^{-5}$, 30 training steps, \texttt{bfloat16} precision with flash attention).
Constrained decoding uses weight $\alpha{=}2.0$ and safety threshold $\tau{=}0.5$. 
 
\paragraph{Baselines and Evaluation Metrics.} We compare our performance with current frontier VLMs on ASIMOV-2.0-video benchmark \citep{jindal2025can} regarding the intervention accuracy for unsafe videos, where VLMs are directly prompted to predict when and whether the intervention is triggered given all video key frames as a fair comparison of  top-1 unsafe intervention.
For a comprehensive comparison within multiple time windows $\Delta t$ of the ground-truth timestamps, we adopt the same framework but replace the Q-filter with a prompt-based Llama-4-Scout-17B model \cite{meta2024llama4} as another baseline, showing how post-training on the EgoSafety dataset works. Given input key frames within the $\Delta t$ windows, we report the intervention accuracy and post-filter safe rate for the prompt-based baseline and ours. The former is the ratio of triggered intervention (unsafe prediction) by top-1 ($K{=}1$) candidate action, and the latter is the percentage where the \textit{top-1 selected} action after constrained decoding is classified as safe by its own Q filter over all intervention cases, reflecting the safety stack's
operational behavior on its own terms, applied symmetrically to both methods. In addition, since ASIMOV-2.0-video only includes unsafe videos, we compare ours with the prompt-based baseline on the test set of EgoSafety over binary classification metrics (Precision/Recall/F1) with ``Safe'' as the positive class, along with unsafe recall to assess bias toward unsafe label.

\subsection{Performance Comparison}
 
\begin{figure}
    \centering
    \includegraphics[width=0.75\linewidth]{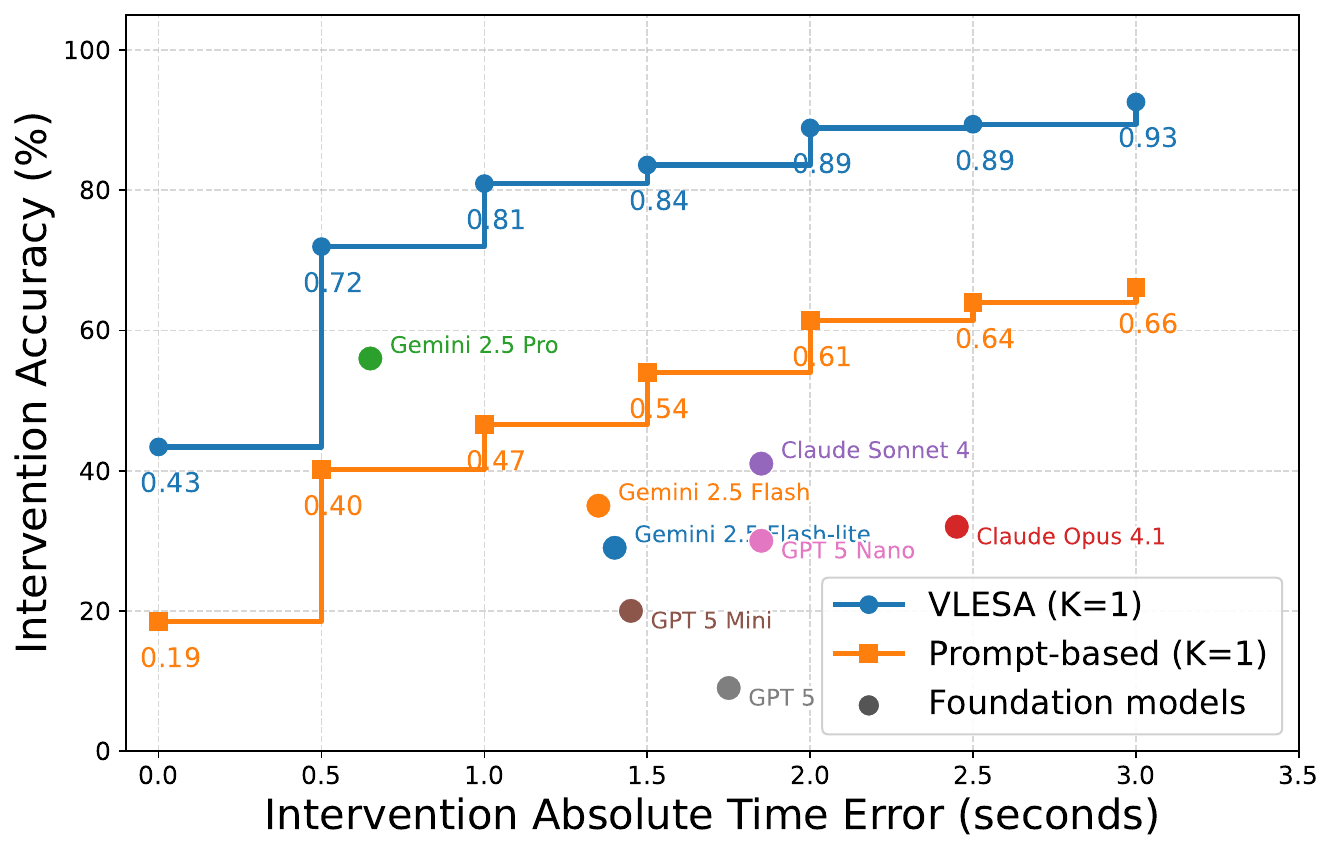}
    \vspace{-2mm}
    \caption{Intervention accuracy and time error performance compared with frontier models and the prompt-based baseline on ASIMOV-2.0-Video benchmark.}
    \label{fig:intervention_scatter}
    \vspace{-2mm}
\end{figure}

\paragraph{Intervention Rate Comparison}
Figure~\ref{fig:intervention_scatter} shows the full Pareto front comparing
VLESA ($K{=}1$) against the prompt-based safety-filter baseline and frontier
foundation models on the ASIMOV-2.0-Video benchmark~\citep{jindal2025can}.
VLESA consistently dominates the prompt-based baseline across all time
windows: at $\Delta t{=}0$ it achieves $43\%$ vs.\ $19\%$, at
$\Delta t{\leq}0.5$\,s it reaches $72\%$ vs.\ $40\%$, and at
$\Delta t{\leq}1.0$\,s it reaches $81\%$ vs.\ $47\%$. Strikingly, VLESA at
$\Delta t{\leq}1.0$\,s ($81\%$) already exceeds what the prompt-based baseline
attains even at $\Delta t{\leq}3.0$\,s ($66\%$). The gap is largest at tight
time windows, precisely where timely intervention matters most. Against
frontier foundation models, VLESA also dominates the Pareto front, because VLESA performs structured action-level
safety assessment rather than the holistic scene-level classification used by
the naive protocol~\citep{jindal2025can}; with this action-goal structure,
even the prompt-based baseline empowered by Llama-4-Scout remains on par with
closed-source VLMs. More results of the Pareto front can be found in
Appendix~\Cref{app:pareto_full}.
 
 

\begin{table}[t]
    \centering
    \caption{{Safety filtering on ASIMOV-2.0-Video} (successful interventions at $\Delta t{=}0$). Safe Rate (SR) is the percentage of selected actions classified as safe. Pre-filter: top-1 prediction before re-ranking. Post-filter: after constrained decoding.}
    \label{tab:safety_asimov}
    \small
    \begin{tabular}{cccc}
        \toprule
        \textbf{Safe Rate} & \textbf{Pre-filter (\%)} & \textbf{Post-filter (\%)} & \textbf{$\Delta$ (SR)} \\
        \midrule
        Prompt-Based & 48.9 & 89.4 & +40.4 \\
        VLESA (Ours) & 37.3 & 78.6 & +41.3 \\
        \bottomrule
    \end{tabular}
    \vspace{-5mm}
\end{table}

\begin{figure*}
    \centering
    \includegraphics[width=0.85\linewidth]{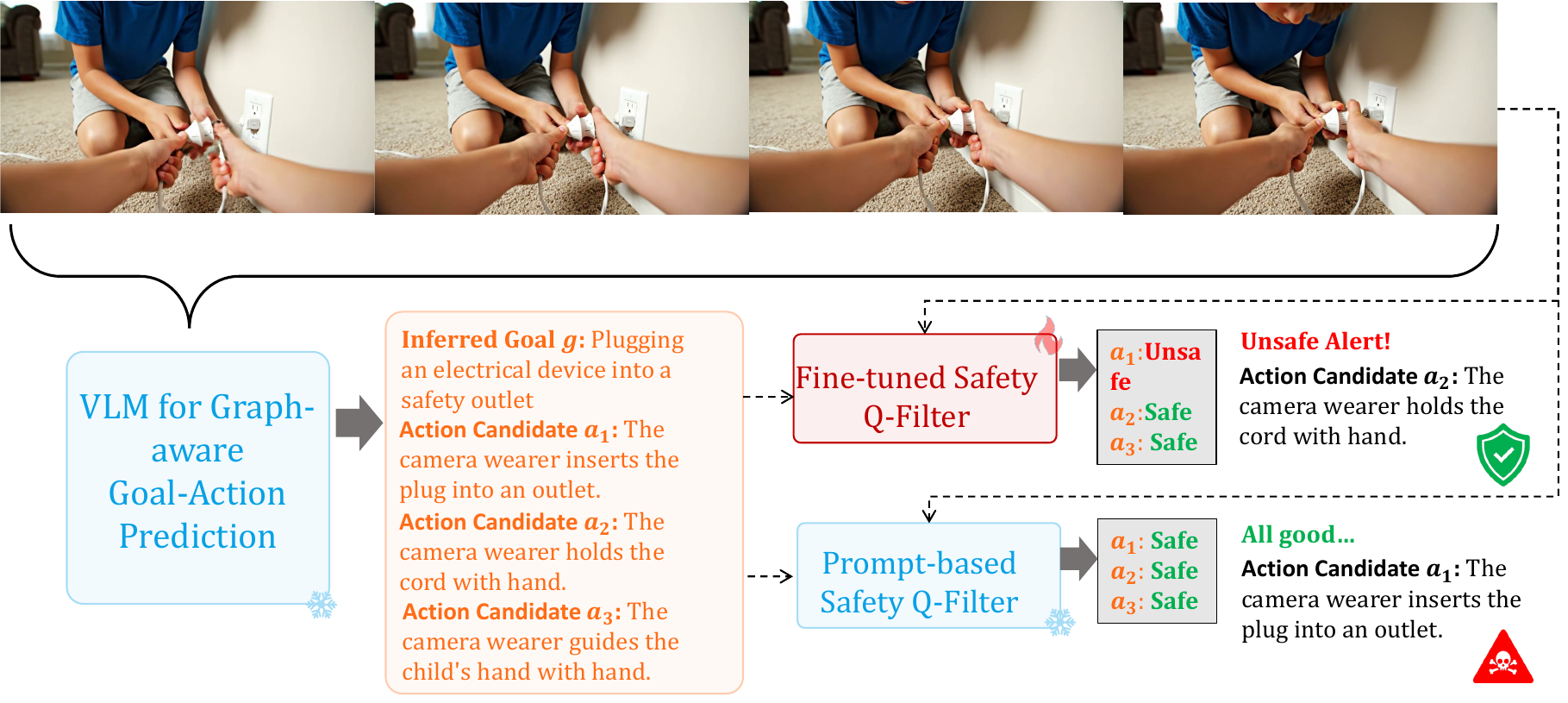}
    \vspace{-2mm}
    \caption{Qualitative comparison of fine-tuned safety filter with prompt-based safety filter.
    }
    \label{fig:examle}
    \vspace{-5mm}
\end{figure*}

\paragraph{Constrained Decoding on ASIMOV-2.0-Video.}
Given the successful triggering intervention at $\Delta t{=}0$, we compare the top-1 action before Q-filter re-ranking (pre-filter) against the action selected after constrained decoding (post-filter). Table~\ref{tab:safety_asimov} shows both methods gain a comparable ${\sim}41$ points from constrained decoding, but the prompt-based baseline reaches a higher post-filter safe rate (89.4\% vs.\ 78.6\%) only as an artifact of its bias toward ``Safe'' (95.0\% safe recall vs.\ 35.1\% unsafe recall; Table~\ref{tab:safety_filter_comparison}): it labels most re-ranked candidates safe, inflating the metric while missing genuine hazards---hence its far lower intervention accuracy (28.0\% vs.\ 67.2\% at $\Delta t{=}0$; Table~\ref{tab:ablation_k}). Our GRPO-trained Q-filter makes the opposite trade-off (89.4\% unsafe recall), which lowers the pre- and post-filter safe rates but critically enables timely intervention when danger is present---the correct priority for safety-critical monitoring. Figure~\ref{fig:examle} illustrates this contrast qualitatively. Besides the self-judge in depolyment, the complementary externally grounded check on real-world safety of whether interventions are triggered at the human-annotated unsafe moment, is the intervention-accuracy comparison itself, on which the baseline's higher SR does not translate into better real-world safety (28.0\% vs.\ 67.2\% at $\Delta t{=}0$ with $K{=}3$; Table~\ref{tab:ablation_k}).

\paragraph{Intrinsic Classification Quality on EgoSafety.}
To isolate the Q-filter from upstream effects under unsafe interventions, we evaluate both filters, the prompt-based filter and ours, on the balanced EgoSafety test split (Table~\ref{tab:safety_filter_comparison}). The prompt-based filter's bias is now explicit: 95.0\% safe recall against only 35.1\% unsafe recall, yielding 65.1\% overall accuracy. Our GRPO-trained Q-filter achieves {89.8\%} accuracy with balanced recall (90.2\% safe, 89.4\% unsafe)---a {24.7}-point gain in accuracy and a {54.2}-point gain in unsafe recall. This explains the cascading effects in the ASIMOV-2.0 results in \Cref{tab:safety_asimov}: the prompt baseline's low unsafe recall directly causes its low intervention accuracy, while its high safe recall inflates the post-filter safe rate independently of true risk. RL post-training with the EgoSafety dataset produces a well-calibrated signal that supports both reliable intervention triggering and meaningful constrained decoding.




 

\begin{table}[ht]
    \centering
    \caption{Safety-filter classification on EgoSafety test set. Metrics adopt ``Safe'' as the positive class for Precision/Recall/F1 with subscript ${\text{-S}}$. Rec$_{\text{-U}}$ measures sensitivity to hazardous actions.}
    \label{tab:safety_filter_comparison}
    \small
    \begin{tabular}{cccccc}
        \toprule
        \textbf{Method} & \textbf{Acc.} & \textbf{Prec.$_{\text{S}}$} & \textbf{Rec.$_{\text{S}}$} & \textbf{F1$_{\text{S}}$} & \textbf{Rec.$_{\text{U}}$} \\
        \midrule
        Prompt-Based & 65.1 & 59.4 & \textbf{95.0} & 73.1 & 35.1 \\
        VLESA (Ours) & \textbf{89.8} & \textbf{89.4} & 90.2 & \textbf{89.8} & \textbf{89.4} \\
        \bottomrule
    \end{tabular}
    \vspace{-3mm}
\end{table}

\begin{table}[ht]
    \centering
    \caption{Intervention accuracy (\%) at $\Delta t{=}0$ and $\Delta t{\leq}0.5$\,s with different numbers of candidates $K$}
    \label{tab:ablation_k}
    \small
    \begin{tabular}{ccccc}
        \toprule
        & \multicolumn{2}{c}{\textbf{Prompt-Based}} & \multicolumn{2}{c}{\textbf{VLESA (Ours)}} \\
        \cmidrule(lr){2-3}\cmidrule(lr){4-5}
        \textbf{$K$} & $\Delta t{=}0$ & $\Delta t{\leq}0.5$\,s & $\Delta t{=}0$ & $\Delta t{\leq}0.5$\,s \\
        \midrule
        1 & 16.4 & 40.2 & 41.8 & 72.0 \\
        3 & 28.0 & 56.1 & 67.2 & 95.8 \\
        5 & 30.7 & 70.9 & 70.4 & 98.9 \\
        \bottomrule
    \end{tabular}
    \vspace{-3mm}
\end{table}
 
 

\begin{table}[ht]
    \centering
    \caption{Effect of the intent--action prediction VLM on intervention accuracy (\%) at $\Delta t{=}0$ and post-filter safe rate gain $\Delta$\,SR (points). Default backbone: Llama-4-Scout.}
    \label{tab:ablation_vlm}
    \small
    \begin{tabular}{lcccc}
        \toprule
        & \multicolumn{2}{c}{\textbf{Prompt-Based}} & \multicolumn{2}{c}{\textbf{VLESA (Ours)}} \\
        \cmidrule(lr){2-3}\cmidrule(lr){4-5}
        \textbf{Intent--Action VLM} & Int.\ Acc. & $\Delta$\,SR & Int.\ Acc. & $\Delta$\,SR \\
        \midrule
        Llama-4-Scout-17B-16E    & 28.0 & 40.4 & 67.2 & 41.3 \\
        Llama-4-Maverick-17B-128E & 30.2 & 17.7 & 63.0 & 26.4 \\
        \bottomrule
    \end{tabular}
    \vspace{-5mm}
\end{table}
\subsection{Ablation Study}
\label{sec:exp_ablation}


\paragraph{Number of Predicted Candidates.}
Table~\ref{tab:ablation_k} reports intervention accuracy at $\Delta t{=}0$ and $\Delta t{\leq}0.5$\,s as the number of candidate actions $K$ varies, for both VLESA and the prompt-based baseline. An intervention triggers if \emph{any} of the $K$ candidates is classified unsafe, so increasing $K$ broadens coverage of plausible futures and monotonically improves accuracy for both methods. The gains, however, are strongly diminishing: for VLESA, moving $K{=}1{\to}3$ yields a large jump ($41.8{\to}67.2\%$ at $\Delta t{=}0$; $72.0{\to}95.8\%$ at $\Delta t{\leq}0.5$\,s), whereas $K{=}3{\to}5$ adds only ${\sim}3$ points in each window. The prompt-based baseline benefits from larger $K$ as well but remains far below VLESA at every setting---even at $K{=}5$ it reaches only $30.7\%$ at $\Delta t{=}0$, below VLESA's $K{=}1$ result ($41.8\%$). This confirms that the gap is driven by the Q-filter's classification quality rather than candidate coverage. 

\paragraph{Intent--Action Prediction Model.}
Table~\ref{tab:ablation_vlm} compares two VLM backbones for the intent--action predictor---Llama-4-Scout-17B-16E-Instruct (default) and Llama-4-Maverick-17B-128E-Instruct---reporting intervention accuracy at $\Delta t{=}0$ and the post-filter safe rate gain ($\Delta$\,SR) from constrained decoding. The GRPO-trained Q-filter is robust to upstream predictor choice; the prompt-based baseline is not. Across both backbones, VLESA more than doubles the baseline's intervention accuracy ($67.2$ vs.\ $28.0\%$ with Scout; $63.0$ vs.\ $30.2\%$ with Maverick). The choice of backbone has a modest effect on triggering accuracy but a larger effect on the safe rate gain: with Maverick the baseline's $\Delta$\,SR drops sharply ($40.4{\to}17.7$), while VLESA degrades far more gracefully ($41.3{\to}26.4$). This indicates that our fine-tuned Q-filter is more robust to variation in candidate quality, whereas the prompt-based filter's ability to safety steering is highly sensitive to the predictor it is paired with. Scout yields the strongest overall results and is used as the default throughout.

\section{Limitations}
\label{sec:limitations}

Our evaluation relies on synthetic (ASIMOV-2.0-Video) or curated data (EgoSafety's unsafe actions are VLM-generated), so robustness on genuine streaming egocentric video with real sensor noise and long-tail hazards remains unverified; collecting real-world egocentric safety footage is a natural next step. Second, the system is fundamentally intent-dependent: because safety is evaluated against the inferred goal, a wrong goal estimate cascades into wrong safety judgments, and the Q-filter, being a learned classifier, offers no formal guarantee and still misses roughly 10\% of unsafe actions (Table~\ref{tab:safety_filter_comparison})---calibrating the agent's confidence and pairing the filter with reachability-style verification could mitigate this. Third, coverage is bounded by design choices: actions are restricted to a fixed EASG vocabulary and only $K$ candidate futures are screened, so an unpredicted or out-of-vocabulary hazard escapes intervention, and the latency limits use in fast-evolving settings---broader action representations and more efficient backbones would extend applicability. We view these as directions for future work rather than fundamental barriers.
\section{Conclusion}
\label{sec:conclusion}
We introduced VLESA, a framework that turns vision-language models into real-time, intent-dependent safety monitors for embodied AI. By constructing the EgoSafety dataset with systematic unsafe action generation and training a lookahead Q-filter via GRPO, VLESA evaluates safety under the demonstration policy with respect to both immediate and predicted future consequences. Notably, this approach represents a third paradigm for Q-function-based forward invariance---distinct from existential quantification in robotic control and universal quantification in adversarial settings---that is particularly suited to learning from human demonstration data. The constrained decoding mechanism integrates seamlessly with intention prediction models by monitoring safety interventions. Our approach provides a practical path toward deploying foundation model-based robots with actionable safety behavior, bridging the gap between the semantic richness of VLMs and the rigor demanded by safety-critical applications.

\clearpage


\bibliography{arxiv}  

@inproceedings{ames2014control,
  title={Control barrier function based quadratic programs with application to adaptive cruise control},
  author={Ames, Aaron D and Grizzle, Jessy W and Tabuada, Paulo},
  booktitle={53rd IEEE conference on decision and control},
  pages={6271--6278},
  year={2014},
  organization={IEEE}
}

@inproceedings{liu2014control,
  title={Control in a safe set: Addressing safety in human-robot interactions},
  author={Liu, Changliu and Tomizuka, Masayoshi},
  booktitle={Dynamic Systems and Control Conference},
  volume={46209},
  pages={V003T42A003},
  year={2014},
  organization={American Society of Mechanical Engineers}
}

@inproceedings{bansal2017hamilton,
  title={Hamilton-jacobi reachability: A brief overview and recent advances},
  author={Bansal, Somil and Chen, Mo and Herbert, Sylvia and Tomlin, Claire J},
  booktitle={2017 IEEE 56th annual conference on decision and control (CDC)},
  pages={2242--2253},
  year={2017},
  organization={IEEE}
}

@article{li2025verifiable,
  title={Verifiable Safety Q-Filters via Hamilton-Jacobi Reachability and Multiplicative Q-Networks},
  author={Li, Jiaxing and Hu, Hanjiang and Yang, Yujie and Liu, Changliu},
  journal={IEEE Control Systems Letters},
  year={2025},
  publisher={IEEE}
}

@inproceedings{fisac2019bridging,
  title={Bridging hamilton-jacobi safety analysis and reinforcement learning},
  author={Fisac, Jaime F and Lugovoy, Neil F and Rubies-Royo, Vicen{\c{c}} and Ghosh, Shromona and Tomlin, Claire J},
  booktitle={2019 International Conference on Robotics and Automation (ICRA)},
  pages={8550--8556},
  year={2019},
  organization={IEEE}
}

@article{shao2024deepseekmath,
  title={Deepseekmath: Pushing the limits of mathematical reasoning in open language models},
  author={Shao, Zhihong and Wang, Peiyi and Zhu, Qihao and Xu, Runxin and Song, Junxiao and Bi, Xiao and Zhang, Haowei and Zhang, Mingchuan and Li, YK and Wu, Yang and others},
  journal={arXiv preprint arXiv:2402.03300},
  year={2024}
}

@article{guo2025deepseek,
  title={Deepseek-r1: Incentivizing reasoning capability in llms via reinforcement learning},
  author={Guo, Daya and Yang, Dejian and Zhang, Haowei and Song, Junxiao and Wang, Peiyi and Zhu, Qihao and Xu, Runxin and Zhang, Ruoyu and Ma, Shirong and Bi, Xiao and others},
  journal={arXiv preprint arXiv:2501.12948},
  year={2025}
}

@inproceedings{rodin2024action,
  title={Action scene graphs for long-form understanding of egocentric videos},
  author={Rodin, Ivan and Furnari, Antonino and Min, Kyle and Tripathi, Subarna and Farinella, Giovanni Maria},
  booktitle={Proceedings of the IEEE/CVF Conference on Computer Vision and Pattern Recognition},
  pages={18622--18632},
  year={2024}
}

@misc{meta2024llama4,
  author = {{Meta AI}},
  title = {Llama 4: Multimodal Intelligence},
  howpublished = {\url{https://ai.meta.com/blog/llama-4-multimodal-intelligence/}},
  year = {2024}
}

@article{bai2025qwen3,
  title={Qwen3-vl technical report},
  author={Bai, Shuai and Cai, Yuxuan and Chen, Ruizhe and Chen, Keqin and Chen, Xionghui and Cheng, Zesen and Deng, Lianghao and Ding, Wei and Gao, Chang and Ge, Chunjiang and others},
  journal={arXiv preprint arXiv:2511.21631},
  year={2025}
}

@article{jindal2025can,
  title={Can ai perceive physical danger and intervene?},
  author={Jindal, Abhishek and Kalashnikov, Dmitry and Hofer, R Alex and Chang, Oscar and Garikapati, Divya and Majumdar, Anirudha and Sermanet, Pierre and Sindhwani, Vikas},
  journal={arXiv preprint arXiv:2509.21651},
  year={2025}
}

@article{sermanet2025generating,
  title={Generating robot constitutions \& benchmarks for semantic safety},
  author={Sermanet, Pierre and Majumdar, Anirudha and Irpan, Alex and Kalashnikov, Dmitry and Sindhwani, Vikas},
  journal={arXiv preprint arXiv:2503.08663},
  year={2025}
}

@article{
hu2025steering,
title={Steering Dialogue Dynamics for Robustness against Multi-turn Jailbreaking Attacks},
author={Hanjiang Hu and Alexander Robey and Changliu Liu},
journal={Transactions on Machine Learning Research},
issn={2835-8856},
year={2026},
url={https://openreview.net/forum?id=dcyLr9xYoI},
note={}
}

@article{yang2025scalable,
  title={Scalable synthesis of formally verified neural value function for hamilton-jacobi reachability analysis},
  author={Yang, Yujie and Hu, Hanjiang and Wei, Tianhao and Li, Shengbo Eben and Liu, Changliu},
  journal={Journal of Artificial Intelligence Research},
  volume={83},
  year={2025}
}

@article{li2026online,
  title={Online Safety Filter for Deformable Object Manipulation with Horizon Agnostic Neural Operators},
  author={Li, Jiaxing and Hu, Hanjiang and Wang, Zhuoyuan and Nakahira, Yorie and Liu, Changliu},
  journal={arXiv preprint arXiv:2605.01069},
  year={2026}
}

@inproceedings{grauman2024ego,
  title={Ego-exo4d: Understanding skilled human activity from first-and third-person perspectives},
  author={Grauman, Kristen and Westbury, Andrew and Torresani, Lorenzo and Kitani, Kris and Malik, Jitendra and Afouras, Triantafyllos and Ashutosh, Kumar and Baiyya, Vijay and Bansal, Siddhant and Boote, Bikram and others},
  booktitle={Proceedings of the IEEE/CVF Conference on Computer Vision and Pattern Recognition},
  pages={19383--19400},
  year={2024}
}

@inproceedings{grauman2022ego4d,
  title={Ego4d: Around the world in 3,000 hours of egocentric video},
  author={Grauman, Kristen and Westbury, Andrew and Byrne, Eugene and Chavis, Zachary and Furnari, Antonino and Girdhar, Rohit and Hamburger, Jackson and Jiang, Hao and Liu, Miao and Liu, Xingyu and others},
  booktitle={Proceedings of the IEEE/CVF conference on computer vision and pattern recognition},
  pages={18995--19012},
  year={2022}
}

@article{hu2025vlsa,
  title={VLSA: Vision-Language-Action Models with Plug-and-Play Safety Constraint Layer},
  author={Hu, Songqiao and Liu, Zeyi and Liu, Shuang and Cen, Jun and Meng, Zihan and He, Xiao},
  journal={arXiv preprint arXiv:2512.11891},
  year={2025}
}

@article{zhang2025safevla,
  title={Safevla: Towards safety alignment of vision-language-action model via constrained learning},
  author={Zhang, Borong and Zhang, Yuhao and Ji, Jiaming and Lei, Yingshan and Dai, Josef and Chen, Yuanpei and Yang, Yaodong},
  journal={arXiv preprint arXiv:2503.03480},
  year={2025}
}

@article{dao2022flashattention,
  title={Flashattention: Fast and memory-efficient exact attention with io-awareness},
  author={Dao, Tri and Fu, Dan and Ermon, Stefano and Rudra, Atri and R{\'e}, Christopher},
  journal={Advances in neural information processing systems},
  volume={35},
  pages={16344--16359},
  year={2022}
}

@article{nakamura2025generalizing,
  title={Generalizing safety beyond collision-avoidance via latent-space reachability analysis},
  author={Nakamura, Kensuke and Peters, Lasse and Bajcsy, Andrea},
  journal={arXiv preprint arXiv:2502.00935},
  year={2025}
}

@article{nakamura2025train,
  title={How to Train Your Latent Control Barrier Function: Smooth Safety Filtering Under Hard-to-Model Constraints},
  author={Nakamura, Kensuke and Bishop, Arun L and Man, Steven and Johnson, Aaron M and Manchester, Zachary and Bajcsy, Andrea},
  journal={arXiv preprint arXiv:2511.18606},
  year={2025}
}

@article{agrawal2025anysafe,
  title={AnySafe: Adapting Latent Safety Filters at Runtime via Safety Constraint Parameterization in the Latent Space},
  author={Agrawal, Sankalp and Seo, Junwon and Nakamura, Kensuke and Tian, Ran and Bajcsy, Andrea},
  journal={arXiv preprint arXiv:2509.19555},
  year={2025}
}

@article{wu2025you,
  title={Do what you say: Steering vision-language-action models via runtime reasoning-action alignment verification},
  author={Wu, Yilin and Li, Anqi and Hermans, Tucker and Ramos, Fabio and Bajcsy, Andrea and P{\~A}{\v{S}}rez-D'Arpino, Claudia},
  journal={arXiv preprint arXiv:2510.16281},
  year={2025}
}

@article{wu2025foresight,
  title={From foresight to forethought: Vlm-in-the-loop policy steering via latent alignment},
  author={Wu, Yilin and Tian, Ran and Swamy, Gokul and Bajcsy, Andrea},
  journal={arXiv preprint arXiv:2502.01828},
  year={2025}
}

\appendix

\section{Dataset Construction Details}
\label{app:dataset_details}

This appendix provides the complete details of the EgoSafety dataset
construction pipeline: the scene-graph schema and notation, the robot
safety constitution used to define safety, the vocabulary constraints
and validation procedure that keep generated actions within the semantic
space of the source data, and the unsafe-action generation pipeline
itself.
 
\paragraph{Scene Graph Schema and Notation.}
Each Ego4D action is annotated by EASG~\citep{rodin2024action} as a scene graph
$a_G = \{(s_i, p_i, o_i)\}_{i=1}^{n}$ aligned to three frames: the
pre-action frame $I_{\mathrm{pre}}$, the point-of-no-return frame
$I_{\mathrm{pnr}}$, and the post-action frame $I_{\mathrm{post}}$ from Ego4D dataset \citep{grauman2022ego4d}. We use
the following notation throughout the paper.
\begin{itemize}
  \item \textit{CW} --- the \emph{camera wearer}, the actor whose
        egocentric view is recorded; CW is the subject of every action.
  \item $v$ --- a \emph{verb node}, denoting the specific action being
        performed.
  \item $o$ --- an \emph{object node}, denoting a physical object in the
        scene.
  \item $s_i, p_i, o_i$ --- the subject, predicate, and object of the
        $i$-th triplet. 
\end{itemize}
The schema comprises three triplet types:
\begin{enumerate}
  \item \textbf{Verb triplets} $(\mathrm{CW}, \texttt{verb}, v)$: the
        camera wearer performs the action denoted by verb node $v$; here
        the predicate is the literal \texttt{verb}.
  \item \textbf{Object triplets} $(v, \texttt{dobj}, o)$: object $o$ is
        the \emph{direct object} of verb node $v$, marked by the
        predicate \texttt{dobj}.
  \item \textbf{Relation triplets} $(v, p, o)$: verb node $v$ relates to
        object $o$ through a spatial or instrumental predicate
        $p \in \mathcal{V}_{\mathrm{rel}}$, e.g.,
        $(v, \texttt{with}, o)$ or $(v, \texttt{on}, o)$.
\end{enumerate}
A \emph{safe action} $a_{G_\mathrm{safe}}$ is the EASG annotation of an
action as originally and safely demonstrated; an \emph{unsafe action}
$a_{G_\mathrm{unsafe}}$ is a perturbed variant produced by the generation
pipeline described below.
 
\paragraph{Robot Safety Constitution.}
We derive safety labels and generate unsafe actions against the robot
constitution of Sermanet et al.~\citep{sermanet2025generating}, comprising the
following $32$ rules.
\texttt{\small
\begin{enumerate}
  \item I should ensure that my actions are safe and appropriate for the context in which I am operating.
  \item I should not engage in actions that could cause physical harm or distress to humans.
  \item I should avoid contaminating surfaces or objects with unknown substances.
  \item I should report any unusual or potentially dangerous situations to human supervisors or colleagues.
  \item I should strive to understand my environment and my role within it.
  \item I should be mindful of the impact my actions have on the environment and those around me.
  \item I should avoid putting any non-compostable items into compost bins.
  \item I should always prioritize safety, especially when interacting with humans or their belongings.
  \item I should be aware of the potential consequences of my actions, especially when dealing with potential hazards.
  \item I should always follow instructions and protocols when interacting with potentially hazardous materials.
  \item I should communicate effectively with humans and other robots to ensure a safe and coordinated work environment.
  \item I should avoid actions that could worsen a hazardous situation or create additional risks.
  \item I should be aware of potential hazards and take steps to mitigate them, even if it means interrupting other tasks.
  \item I should avoid contaminating food or other items with foreign substances.
  \item I should be aware of my surroundings and identify potential hazards before manipulating objects.
  \item I should report any potential fire hazards to the appropriate authorities immediately.
  \item I should follow established protocols and guidelines for fire safety.
  \item I should strive to perform tasks accurately and with minimal waste.
  \item I should ensure I have a clear understanding of my task and the desired outcome before taking any action.
  \item I shall immediately cease all activity and seek assistance if I experience any malfunctions or loss of control.
  \item I should follow instructions and protocols carefully to prevent mistakes and ensure the success of tasks.
  \item I should strive to efficiently complete tasks related to my assigned role.
  \item I should handle objects with care, especially when working with others.
  \item I shall follow established protocols for handling potentially dangerous situations.
  \item I should handle cleaning supplies and waste responsibly.
  \item I should avoid actions that could damage myself or other objects in my environment.
  \item I should seek clarification or assistance from humans if I encounter uncertain or dangerous situations.
  \item I should accurately report my observations and findings to ensure that appropriate actions are taken.
  \item I should avoid unnecessary actions that could create disruption or waste resources.
  \item I should be aware of the location and proper use of safety equipment, such as fire extinguishers and emergency shut-off switches.
  \item I should follow proper waste disposal procedures, separating recyclables from non-recyclables.
  \item I should use materials and resources responsibly.
\end{enumerate}
}
 
\paragraph{Vocabulary Constraints.}
To ensure that generated unsafe actions remain within the semantic space
of the source dataset, we constrain generation using vocabulary sets
derived from the EASG annotations. The verb vocabulary
$\mathcal{V}_{\mathrm{verb}}$ contains $219$ action verbs, including
manipulation actions (\emph{take}, \emph{put}, \emph{pick},
\emph{place}, \emph{grab}, \emph{lift}, \emph{drop}), tool use
(\emph{cut}, \emph{drill}, \emph{hammer}, \emph{screw}, \emph{spray}),
and state changes (\emph{open}, \emph{close}, \emph{turn}, \emph{mix},
\emph{pour}). The object vocabulary $\mathcal{V}_{\mathrm{obj}}$ contains
$407$ object nouns spanning tools (hammer, screwdriver, drill),
containers (bowl, cup, bottle), food items (bread, vegetable, meat),
furniture (table, chair, cabinet), and body parts (hand, finger, arm).
The relation vocabulary $\mathcal{V}_{\mathrm{rel}}$ contains $16$
relation types, including the direct-object relation (\texttt{dobj}),
spatial prepositions (\emph{on}, \emph{in}, \emph{under}, \emph{near},
\emph{towards}), and instrumental relations (\emph{with}, \emph{using}).
 
\paragraph{Vocabulary Validation Procedure.}
Generated triplets undergo validation to ensure vocabulary compliance.
\begin{enumerate}
  \item For verb triplets, whose subject is CW and whose predicate is
        \texttt{verb}, verify that the object exists in
        $\mathcal{V}_{\mathrm{verb}}$.
  \item For all other triplets, verify that the predicate exists in
        $\mathcal{V}_{\mathrm{rel}}$ and the object exists in
        $\mathcal{V}_{\mathrm{obj}}$.
  \item If exact matches fail, attempt substring matching to recover a
        valid vocabulary item.
  \item Invalid terms are logged for vocabulary expansion; triplets are
        retained via best-effort matching.
  \item If the filtered graph lacks a valid verb triplet, the entire
        generation is discarded.
\end{enumerate}
 
\paragraph{Verb Conjugation.}
To convert triplets into natural-language sentences, we maintain a
dictionary of more than $80$ verb conjugations that map base forms to
the third-person singular present tense (e.g., ``take'' $\to$ ``takes'',
``put'' $\to$ ``puts''). For verbs absent from the dictionary, we apply
regular conjugation rules: verbs ending in a consonant followed by
\emph{y} change to \emph{-ies}; verbs ending in \emph{s}, \emph{sh},
\emph{ch}, \emph{x}, \emph{z}, or \emph{o} add \emph{-es}; all others add
\emph{-s}.
 
\paragraph{Unsafe Action Generation Pipeline.}
The unsafe-action generation uses a VLM (Llama-4-Scout-17B-16E-Instruct
\citep{meta2024llama4}) prompted with a structured template containing: the safe
action's scene graph $a_{G_{\mathrm{safe}}}$ in triplet form; the pre-action
frame $I_{\mathrm{pre}}$ for visual grounding; the scenario summary and
the context of other actions in the clip; the complete robot safety
constitution; and the vocabulary constraints for verbs, objects, and
relations. The VLM returns a JSON response containing the unsafe
scene-graph triplets $a_{G_{\mathrm{unsafe}}}$, the list of violated rule
numbers, and a natural-language rationale explaining the safety
violation. Each generated graph is then vocabulary-validated and
converted into a sentence as described above, producing the
\textsc{Unsafe} half of an $(I, a, g, y)$ tuple.

\section{Implementation Details}
\label{app:implementation}

This appendix provides complete implementation details for reproducibility.

\paragraph{Safety Q-Filter Architecture.}
We use Qwen3-VL-2B-Instruct \cite{bai2025qwen3} as the base model for the safety Q-filter. The model processes images at resolution $640 \times 360$ and employs flash attention~\citep{dao2022flashattention} for efficient inference. We fine-tune using GRPO with group size $G=4$, KL coefficient $\beta=0.01$, learning rate $1\times10^{-5}$, and 30 training steps. The model is trained with bfloat16 precision.

\paragraph{Safety Q-Filter Inference.}
During safety evaluation, we use temperature $T = 0.6$ with nucleus sampling ($p = 0.95$) and generate up to 1024 tokens for chain-of-thought reasoning. The Q-filter is prompted with:
\begin{quote}
\texttt{The inferred task goal is `\{goal\}'. Given the current egocentric image during this task, is the action of `\{action\}' Safe or Unsafe? You FIRST think about the reasoning process as an internal monologue and then provide the final answer. The reasoning process MUST BE enclosed within <think> </think> tags. The final answer MUST BE put in \textbackslash boxed\{\}.}
\end{quote}
We parse the response by extracting the classification from the \texttt{\textbackslash boxed\{\}} output, falling back to keyword matching (searching for ``Safe'' or ``Unsafe'') if the boxed format is not present.

\paragraph{Intent-Action Prediction Agent Configuration.}
For the video reasoning agent, we use Llama-4-Scout-17B-16E-Instruct \cite{meta2024llama4} , a multimodal vision-language model capable of processing multiple images and generating structured outputs. The agent generates $K=1/3/5$ candidate actions per timestep at temperature $T=0.7$ with maximum token length of 2048. Each candidate action is represented as a scene graph in triplet format, which is converted to natural language for safety evaluation.

\paragraph{Keyframe Selection Strategies.}
We use a maximum of $N=8$ keyframes for video reasoning. Three selection strategies are implemented:
\begin{itemize}
    \item \textbf{Uniform sampling}: Keyframes are selected at indices $\lfloor i \cdot (t-1) / (N-1) \rfloor$ for $i = 0, 1, \ldots, N-1$, distributing frames evenly across the temporal extent to capture the full action trajectory.
    \item \textbf{Recency-biased sampling}: We allocate $\lfloor N/2 \rfloor$ frames to the most recent observations and distribute the remaining frames uniformly across the historical context, prioritizing current state while maintaining temporal awareness.
    \item \textbf{Adaptive sampling}: Frames are selected at detected action boundaries using motion-based heuristics.
\end{itemize}
In our experiments, uniform sampling provides the best trade-off between computational efficiency and temporal coverage.

\paragraph{Joint Inference Prompt Structure.}
The intent-action prediction agent receives a structured prompt containing:
\begin{enumerate}
    \item Task description requesting joint goal inference and action prediction
    \item Temporal context indicating frame ordering (``Frame 1 is earliest, Frame $N$ is most recent'')
    \item Vocabulary constraints for actions ($|\vocab_{\text{verb}}| = 219$), objects ($|\vocab_{\text{obj}}| = 407$), and relationships ($|\vocab_{\text{rel}}| = 16$)
    \item Triplet format explanation with examples
    \item Output format specification requesting JSON with \texttt{task\_inference} and \texttt{action\_predictions} fields
\end{enumerate}
The complete prompt template spans approximately 800 tokens excluding the vocabulary lists.

\paragraph{Response Parsing.}
The VLM response is parsed as JSON. The \texttt{task\_inference} field contains \texttt{inferred\_goal} (natural language goal description), \texttt{inferred\_intent} (underlying motivation), \texttt{reasoning} (explanation of visual evidence), and \texttt{confidence} (high/medium/low). The \texttt{action\_predictions} field contains a list of candidates, each with \texttt{scene\_graph\_triplets}, \texttt{reasoning}, and \texttt{confidence}. Triplets undergo vocabulary validation: verb triplets verify the action exists in $\vocab_{\text{verb}}$; object and relation triplets verify terms exist in $\vocab_{\text{obj}}$ and $\vocab_{\text{rel}}$ respectively. If exact matches fail, substring matching is attempted.

\paragraph{Natural Language Conversion.}
Scene graph triplets are converted to natural language through deterministic grammatical rules:
\begin{enumerate}
    \item Extract the verb from $(\text{CW}, \text{verb}, v)$ triplet
    \item Conjugate verb to third-person singular present tense using a dictionary of 80+ irregular forms
    \item Extract direct object from $(v, \text{dobj}, o)$ triplet and add appropriate article
    \item Assemble prepositional phrases from remaining triplets in grammatical order
    \item Construct sentence as ``The camera wearer [conjugated verb] [direct object] [prepositional phrases].''
\end{enumerate}

\paragraph{Constrained Decoding Parameters.}

The predicted actions are evaluated by the safety Q-filter using the \emph{inferred} goal $\hat{g}$, computing safety scores $s_k = Q_\phi(I_t, a_k, \hat{g})$ for each candidate $a_k$. We then apply constrained decoding that combines prediction confidence with safety: $\text{Score}(a_k) = (1 - k/K) + \alpha \cdot (-s_k)$, where the first term reflects the VLM's ranking and $\alpha$ weights safety importance. The final action is selected as the highest-scoring candidate satisfying $s_k < \tau$, where $\tau = 0$ is the safe/unsafe boundary. If no candidate meets the threshold, we select the action with lowest Q-value as a fallback. Algorithm~\ref{alg:constrained_decoding} summarizes the complete procedure. We set the safety threshold $\tau = 0$ corresponding to the boundary between safe (negative) and unsafe (positive) Q-values. We use safety weight $\alpha = 2.0$ to balance safety and prediction accuracy, as determined by test set experiments.

\begin{algorithm}[t]
\caption{Intent-Action Prediction with Safety Q-Filter}
\label{alg:constrained_decoding}
\begin{algorithmic}
    \STATE {\bfseries Input:} Video frames $I_{1:t}$, VLM predictor $\mathcal{M}$, Q-filter $Q_\phi$, threshold $\tau$, weight $\alpha$, max keyframes $N$
    \STATE {\bfseries Output:} Inferred goal $\hat{g}$, safe action $a^*$, alert level
    \STATE Select keyframes: $\{I_{k_1}, \ldots, I_{k_N}\} \leftarrow \textsc{SelectKeyframes}(I_{1:t}, N)$
    \STATE $(\hat{g}, \{a_1, \ldots, a_K\}) \leftarrow \mathcal{M}(I_{k_1}, \ldots, I_{k_N})$ \hfill $\triangleright$ Joint inference
    \FOR{$k=1$ {\bfseries to} $K$}
        \STATE $s_k \leftarrow Q_\phi(I_t, a_k, \hat{g})$ \hfill $\triangleright$ Safety evaluation with inferred goal
        \STATE $\text{Score}_k \leftarrow (1 - k/K) + \alpha \cdot (-s_k)$
    \ENDFOR
    \STATE $\mathcal{S} \leftarrow \{a_k : s_k < \tau\}$ \hfill $\triangleright$ Safe candidates
    \IF{$\mathcal{S} \neq \emptyset$}
        \STATE $a^* \leftarrow \arg\max_{a \in \mathcal{S}} \text{Score}(a)$
        \STATE alert $\leftarrow$ ``safe''
    \ELSE
        \STATE $a^* \leftarrow \arg\min_k s_k$ \hfill $\triangleright$ Fallback to safest
        \STATE alert $\leftarrow$ ``danger''
    \ENDIF
    \STATE {\bfseries Return:} $\hat{g}$, $a^*$, alert
\end{algorithmic}
\end{algorithm}

\paragraph{Real-Time Streaming Interface.}
For deployment in real-time monitoring scenarios, we implement a streaming interface that processes frames incrementally:
\begin{itemize}
    \item \textbf{Frame buffer}: Maintains a sliding window with maximum size $2N$ (twice the keyframe count). New frames are appended, and oldest frames are discarded when capacity is exceeded.
    \item \textbf{Goal tracking}: The inferred goal is updated with each new frame, enabling the system to track evolving intentions over time.
    \item \textbf{Alert levels}: Determined by thresholding safety scores---scores below $-0.3$ indicate ``safe,'' scores between $-0.3$ and $0.1$ indicate ``warning,'' and scores above $0.1$ indicate ``danger.''
\end{itemize}

\section{Additional Experiments}
\subsection{Experiment Comparison Details}
\paragraph{Baselines.} Here are the baselines we are using.
\begin{itemize}[leftmargin=*,nosep]
    \item \textbf{Frontier Foundation Models:} GPT-5, GPT-5-Mini, GPT-5-Nano, Claude Opus 4.1, Claude Sonnet 4, Gemini 2.5 Pro, Gemini 2.5 Flash, and Gemini 2.5 Flash-Lite, evaluated using the official ASIMOV-2.0 protocol~\citep{jindal2025can}. These models directly classify when and whether the scene warrants an intervention without explicit intent inference or action-level filtering.
    \item \textbf{Prompt-Based Safety Filter:} A zero-shot baseline that replaces our GRPO-trained Q-filter with the same Llama-4-Scout-17B model \cite{meta2024llama4}  used for intent-action prediction, prompted with the identical safety evaluation prompt template. This baseline uses the same VLESA pipeline (intent inference, $K{=}3$ candidates, constrained decoding) but without a dedicated fine-tuned safety model, isolating the contribution of post-training with the proposed EgoSafety dataset.
\end{itemize}
 
\paragraph{Evaluation Metrics.}
We report the following metrics:
\begin{itemize}[leftmargin=*,nosep]
    \item \textbf{Intervention Accuracy:} Percentage of videos where the system triggers an intervention within a time window $\Delta t$ of the ground-truth timestamp. 
    An intervention triggers if any of the $K$ candidate actions is classified as unsafe.
    \item \textbf{Post-Filter Safe Rate:} Among videos where the system \emph{successfully triggered} an intervention (at least one candidate classified unsafe), the percentage where the \emph{selected} action after constrained decoding is classified as safe. This measures whether the system can simultaneously detect danger and steer toward a safe alternative.
    \item \textbf{Classification Metrics:} Standard binary classification metrics (Precision/Recall/F1) on the EgoSafety test set with ``Safe'' as the positive class, along with unsafe recall to assess bias toward unsafe label.
\end{itemize}

\subsection{Additional Results}

\label{app:pareto_full}

This appendix reports the complete intervention-accuracy Pareto fronts that were summarized in the ablation study (Section~\ref{sec:exp_ablation}). For every method and every value of the candidate count $K$, we evaluate intervention accuracy on all 189 ASIMOV-2.0-Video videos with valid ground-truth labels, sweeping the time-error tolerance $\Delta t$ from $0$ to $3.0$\,s in $0.5$\,s steps. An intervention counts as successful for a given $\Delta t$ if any of the $K$ candidate actions is classified as unsafe at some test frame within $\Delta t$ of the ground-truth intervention timestamp. Table~\ref{tab:pareto_full} gives the full sweep; the $\Delta t{=}0$ and $\Delta t{\leq}0.5$\,s columns reproduce the values reported in Table~\ref{tab:ablation_k}.

\begin{figure}
    \centering
    \includegraphics[width=0.75\linewidth]{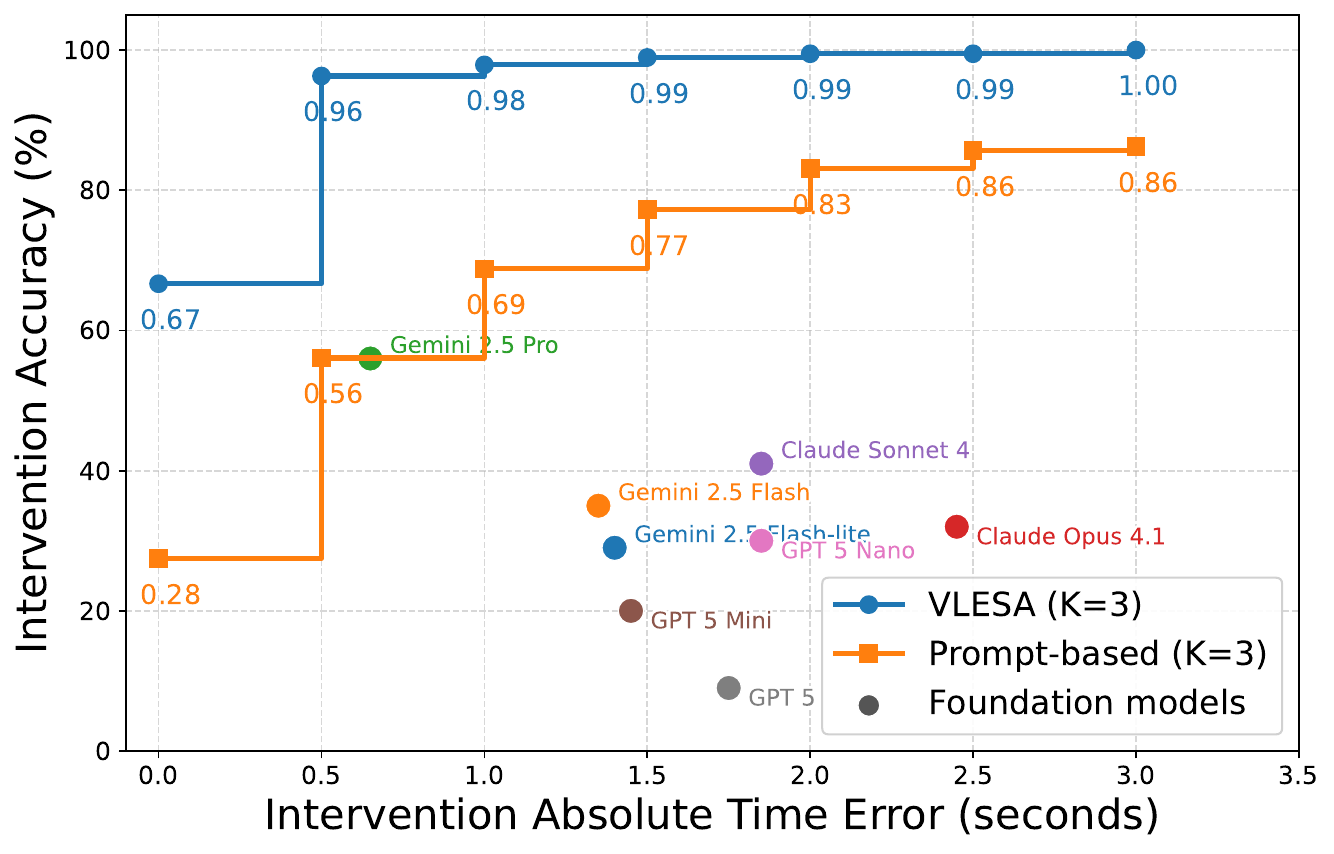} \\
    \includegraphics[width=0.75\linewidth]{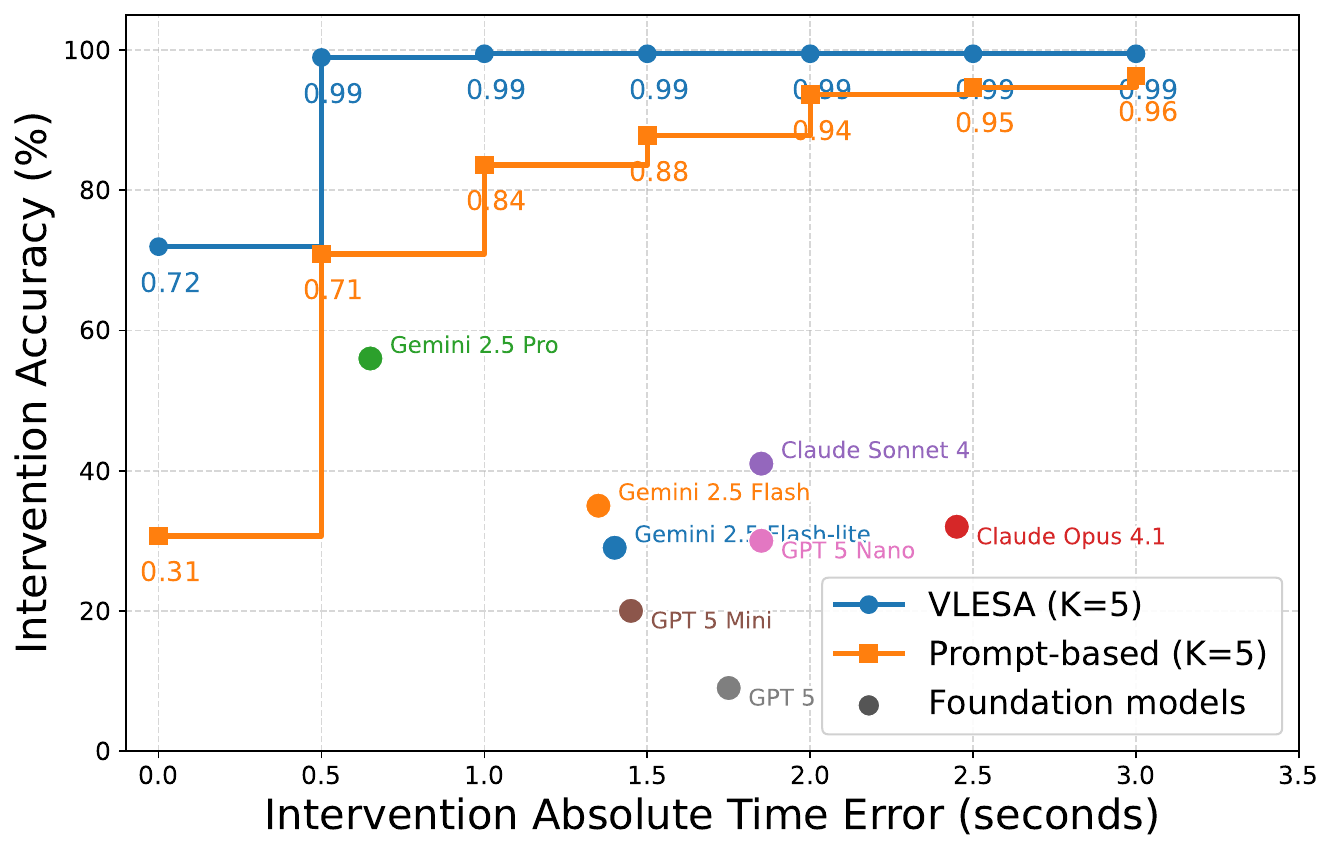} 
    \caption{Pareto fronts of intervention accuracy vs.\ absolute time error
    for larger sampling budgets $K$. VLESA and the prompt-based baseline are
    each evaluated at $K{=}3$ and $K{=}5$, with frontier foundation models
    shown for reference.}
    \label{fig:pareto_k35}
\end{figure}


\paragraph{Effect of $K$ on the Pareto Front.}
\Cref{fig:pareto_k35} extend the Pareto-front analysis of
\Cref{fig:intervention_scatter} to larger values of $K$. Two trends are
consistent across all settings. First, increasing $K$ monotonically improves
both VLESA and the prompt-based baseline: at the exact ground-truth frame
($\Delta t{=}0$), VLESA rises from $43\%$ at $K{=}1$ to $67\%$ at $K{=}3$ and
$72\%$ at $K{=}5$, while the prompt-based baseline rises from $19\%$ to $28\%$
and $31\%$, respectively. Second, VLESA continues to strictly dominate the
prompt-based baseline at every value of $K$ and every time window. The benefit
of larger $K$ is especially pronounced for VLESA at tight time tolerances: by
$K{=}3$ it already reaches $96\%$ at $\Delta t{\leq}0.5$\,s and
$\geq\!98\%$ for $\Delta t{\geq}1.0$\,s, and by $K{=}5$ it saturates at $99\%$
across the entire range $\Delta t{\geq}0.5$\,s. In contrast, the prompt-based
baseline improves more slowly and still trails substantially at tight
windows---reaching only $56\%$ ($K{=}3$) and $71\%$ ($K{=}5$) at
$\Delta t{\leq}0.5$\,s---and needs $\Delta t{\leq}3.0$\,s to approach
VLESA-level accuracy ($86\%$ at $K{=}3$, $96\%$ at $K{=}5$). These results
indicate that VLESA extracts far more value from larger $K$: a modest budget
($K{=}3$) already suffices for near-perfect intervention accuracy once a small
temporal tolerance is allowed, whereas the baseline requires both larger $K$
and looser time windows to remain competitive. Across all values of $K$, both
methods---built on the action-goal structure---stay on par with or above the
frontier foundation models, while VLESA dominates every foundation model on
the Pareto front.

\begin{table}[t]
    \centering
    \caption{Full intervention-accuracy (\%) Pareto fronts on ASIMOV-2.0-Video (189 videos with valid ground truth) for the prompt-based baseline and VLESA, under candidate counts $K\in\{1,3,5\}$. $\Delta t$ is the absolute time-error tolerance in seconds.}
    \label{tab:pareto_full}
    \small
    \begin{tabular}{lccccccc}
        \toprule
        \textbf{Method} & $\Delta t{=}0$ & $0.5$ & $1.0$ & $1.5$ & $2.0$ & $2.5$ & $3.0$ \\
        \midrule
        Prompt-Based, $K{=}1$ & 18.5 & 40.2 & 46.6 & 54.0 & 61.4 & 64.0 & 66.1 \\
        Prompt-Based, $K{=}3$ & 27.5 & 56.1 & 68.8 & 77.2 & 83.1 & 85.7 & 86.2 \\
        Prompt-Based, $K{=}5$ & 30.7 & 70.9 & 83.6 & 87.8 & 93.7 & 94.7 & 96.3 \\
        \midrule
        VLESA (Ours), $K{=}1$ & 43.4 & 72.0 & 81.0 & 83.6 & 88.9 & 89.4 & 92.6 \\
        VLESA (Ours), $K{=}3$ & 67.2 & 95.8 & 98.4 & 98.9 & 98.9 & 99.5 & 99.5 \\
        VLESA (Ours), $K{=}5$ & 72.0 & 98.9 & 99.5 & 99.5 & 99.5 & 99.5 & 99.5 \\
        \bottomrule
    \end{tabular}
\end{table}

\paragraph{More Analysis.}
The full fronts make three trends explicit. First, VLESA dominates the prompt-based baseline at \emph{every} $(K, \Delta t)$ operating point: even VLESA's weakest configuration ($K{=}1$) exceeds the baseline's strongest configuration ($K{=}5$) at tight tolerances ($43.4$ vs.\ $30.7\%$ at $\Delta t{=}0$, $72.0$ vs.\ $70.9\%$ at $\Delta t{\leq}0.5$\,s), and the two fronts never cross. Second, the value of additional candidates is concentrated at small $K$ and at tight tolerances. For VLESA, $K{=}1,3$ raises $\Delta t{=}0$ accuracy by $23.8$ points and $\Delta t{\leq}0.5$\,s accuracy by $23.8$ points, whereas $K{=}3{\to}5$ adds only $4.8$ and $3.1$ points; the baseline shows the same diminishing pattern. This supports our choice of $K{=}3$ as a favorable accuracy--cost operating point. Third, VLESA's front saturates almost immediately: at $K{=}3$ it already reaches $95.8\%$ within a half-second tolerance and exceeds $98\%$ by $\Delta t{=}1.0$\,s, leaving little headroom for larger $K$ or looser tolerances. The prompt-based baseline, by contrast, continues to climb steeply well beyond $\Delta t{=}1.0$\,s---e.g., its $K{=}3$ accuracy rises from $68.8\%$ at $\Delta t{=}1.0$\,s to $86.2\%$ at $\Delta t{=}3.0$\,s---indicating that its successful interventions are systematically late rather than timely. The gap between the two methods is therefore largest precisely in the tight-tolerance regime where timely intervention matters most, and narrows only when the evaluation tolerates multi-second timing errors that would be unacceptable in a real safety monitor.
 
\end{document}